\documentclass[11pt, a4paper]{article}
\usepackage{amsmath, amssymb, amsthm}
\usepackage{mathrsfs}
\usepackage{booktabs}
\usepackage[margin=1in]{geometry}
\usepackage{hyperref} 
\hypersetup{
    colorlinks=true,
    linkcolor=blue,
    citecolor=blue,
    urlcolor=blue,
}

\title{Sheaf-Theoretic Planning: A Categorical Foundation for Resilient Multi-Agent Autonomous Systems}
\author{Manuel Hernández and Eduardo Sánchez-Soto}
\date{Universidad Tecnológica de la Mixteca}

\begin{document}

\maketitle

\begin{abstract}
The challenge of engineering autonomous agents capable of navigating
the stochastic and adversarial nature of the physical world has
historically resided at the intersection of symbolic logic and control
theory. Traditional multi-agent system (MAS) frameworks have relied
heavily on monolithic logical models—primarily variations of the event
calculus and situation calculus—to represent action, change, and
temporal persistence. While these classical systems provide robust
solutions to the frame problem through mechanisms like circumscription
and successor state axioms, they are inherently limited by a
closed-world assumption that fails in the face of unobserved agent
interventions, plan interruptions, and divergent belief-reality
states. The paradigm of Sheaf-Theoretic Planning (STP) emerges as a
transformative alternative, grounding the problem of multi-agent
coordination under the mathematical structures of topos theory and
sheaf semantics. This report provides an exhaustive analysis,
justification, and extension of the STP framework, exploring its
categorical foundations, implementation feasibility, and role in the
future of resilient autonomous systems.
\end{abstract}

\section{The Structural Crisis of Classical Temporal Logic}

Classical temporal reasoning in artificial intelligence is
characterized by an attempt to reduce the complexities of change to a
set of formal axioms governed by an external logical calculus. The
event calculus, first proposed by Kowalski and Sergot, represents the
world through fluents—properties that persist over time—and events
that initiate or terminate these fluents. Persistence is handled by
inertia axioms, where a fluent holds at a time \(t\) if it was
initiated by an earlier event and has not been ``clipped''
(terminated) in the intervening interval. The logic uses
circumscription to solve the frame problem, essentially assuming that
nothing changes unless an event explicitly happens.

However, this reliance on a single, objective timeline breaks down in
open-world scenarios where the ``closed-world'' assumption is
routinely violated. In environments with unobserved agents, a robot
may perceive a world state that contradicts its internal memory. For
instance, a robot \(\mathbb{R}\) that leaves a room to recharge its
battery might return to find that an unobserved robot \(\mathbb{T}\)
has rearranged the environment. In a classical event calculus
framework, this discrepancy is an inconsistency: the logic assumes
that since \(\mathbb{R}\) did not observe an event, no event
occurred. To resolve this, standard systems often require
``extra-logical patches''—computational workarounds that are not
integrated into the core mathematical model of the world.

\begin{table}[htbp]
\centering
\caption{Comparison of Planning Frameworks}
\label{tab:comparison}
\begin{tabular}{@{}p{2.1cm}p{2.8cm}p{3cm}p{3cm}p{2.5cm}@{}}
\toprule
Framework & Temporal Model & Inertia Handling & Open-World Support & Paradigm \\
\midrule
Event Calculus & Linear time & Circumscription & Violation of closed-world assumption & Abductive/ Deductive \\
Situation Calculus & Situations & Successor state axioms & Global transitions struggle with concurrency & Deductive \\
Golog & Basic Action Theories & Deterministic environments & Limited & High-level programming \\
Sheaf-Theoretic Planning & Interval category & Sheaf locality/ gluing & Models multiple inconsistent perspectives & Categorical/ Geometric \\
\bottomrule
\end{tabular}
\end{table}

The limitations extend to the problem of plan interruption and
resource dependency. When an agent's internal state (e.g., low
battery) necessitates an interruption, the classical narrative often
fails to account for the ``blind'' interval. Furthermore, the lack of
a formal mechanism to relate multiple inconsistent viewpoints—such as
an agent's memory, its goal state, and the objective reality—forces
developers to implement complex indexing schemes that do not
scale. Sheaf-theoretic planning addresses these failures by replacing
monolithic logic with a geometric framework where truth is local and
contextual.

\section{Categorical Foundations: The Site of Time}

The fundamental shift in STP is the transition from a point-based timeline to a category of time intervals \(\mathcal{T}\). In this framework, time is not merely a sequence of integers or reals but a poset category where objects are closed intervals \([t_i, t_j]\) and morphisms are interval inclusions \(I_1 \subseteq I_2\). This structure is necessary to capture the duration and overlap of actions, allowing the framework to model temporal granularity and concurrency natively. To imbue this category with causal structure, it is equipped with a Grothendieck topology \(\mathcal{J}\). A Grothendieck topology generalizes the notion of an open cover from topology to arbitrary categories. On the site \((\mathcal{T}, \mathcal{J})\), a ``cover'' of an interval \(I\) is a collection of subintervals whose union is exactly \(I\). This choice formalizes the causal history of the system: the complete history of an interval \(I\) is uniquely determined by the histories of its constituent sub-parts. This is the ``local-to-global'' principle that defines sheaf theory.

A sheaf \(\mathcal{F}\) on this site assigns a set of ``sections''—which we interpret as consistent histories—to each interval \(I\). The sheaf axioms ensure that:
\begin{enumerate}
    \item \textbf{Locality:} If two histories over an interval \(I\) are identical when restricted to every subinterval in a cover of \(I\), then they are identical.
    \item \textbf{Gluing:} If we have a set of compatible local histories (e.g., Robot A's observation of \(I_1\) and Robot B's observation of \(I_2\)), they can be uniquely ``glued'' into a global history over \(I\).
\end{enumerate}

This categorical setup allows for the simultaneous existence of multiple sheaves over the same site. We define \(\mathcal{F}_{\text{World}}\) for objective reality, \(\mathcal{F}_{\text{Mem}}\) for an agent's memory, and \(\mathcal{F}_{\text{Goal}}\) for its target states. The topos \(\mathbf{Sh}(\mathcal{T}, \mathcal{J})\) thus becomes a mathematical universe where multiple ``realities'' are related by natural transformations and pullbacks, providing a rigorous foundation for reasoning about belief discrepancies and unobserved interventions.

\section{Actions as Natural Transformations}

In classical planning, an action is a transition between states. In STP, an action \(\alpha\) is defined as a natural transformation \(\eta_\alpha : \mathcal{F}_{\text{World}} \to \mathcal{F}_{\text{World}}\). A natural transformation is a morphism between functors (in this case, sheaves) that preserves the underlying structure of the category \(\mathcal{T}\). This means that if an action is applied to a history over a long interval \(I\), its effect on any subinterval \(J \subseteq I\) must be consistent with the action applied directly to the history over \(J\).

This naturality condition is essential for ensuring temporal consistency in distributed systems. It generalizes the \textit{initiates} and \textit{terminates} predicates of the event calculus into a geometric framework. When an agent ``moves'' a block, the natural transformation \(\eta_{\text{Move}}\) modifies the stalks—the local state of fluents at each time point—to reflect the new configuration. Because natural transformations compose, a plan becomes a composite transformation in the topos.

\begin{table}[htbp]
\centering
\caption{Categorical Concepts and Planning Interpretation}
\label{tab:categorical}
\begin{tabular}{@{}p{3.7cm}p{4cm}p{7cm}@{}}
\toprule
Categorical Concept & Planning Interpretation & Significance for Resilience \\
\midrule
Natural Transformation & Action that modifies histories consistently & Ensures actions respect temporal inclusions \\
Stalk & State of the world at an instant \(t\) & Allows point-wise evaluation of fluents \\
Restriction Map & Looking at a portion of a history & Models prefix closure and memory \\
Composition & Sequence of actions (a plan) & Facilitates recursive decomposition \\
\bottomrule
\end{tabular}
\end{table}

The power of this approach lies in the separation of the action logic from the global state. In classical systems, an unobserved action creates a contradiction in the global theory. In the sheaf model, an unobserved action is simply a transformation that has been applied to \(\mathcal{F}_{\text{World}}\) but not yet reflected in \(\mathcal{F}_{\text{Mem}}\). The discrepancy is not a logical failure but a geometric distance that can be measured and resolved.

\section{The Geometry of Abduction and Pullbacks}

Abduction—inference to the best explanation—is often the weakest link in autonomous reasoning systems, typically requiring probabilistic heuristics or non-monotonic overrides. STP formalizes abduction as a first-class operation via the pullback construction. A pullback is a limit in category theory that represents the most general way to reconcile two different paths to a common object.

When Robot R returns from recharging and perceives a state \(S_{\text{obs}}\) that contradicts its memory \(S_{\text{mem}}\), it must find an explanation. Let \(\mathcal{H}\) be the sheaf of possible action sequences. We define a mapping \(\Phi: \mathcal{H} \to \mathcal{S}_{\text{space}}\) that takes an action sequence and produces the state resulting from its execution starting at \(S_{\text{mem}}\). We also have a constant mapping \(* \to \mathcal{S}_{\text{space}}\) that picks out the observed state \(S_{\text{obs}}\). The pullback \(P = \mathcal{H} \times_{\mathcal{S}_{\text{space}}} \{*\}\) is the set of all action sequences that transform \(S_{\text{mem}}\) into \(S_{\text{obs}}\).

This construction provides several deep insights:
\begin{itemize}
    \item Abduction is the ``inverse'' of state transition: while transitions move from state to state via actions, abduction moves from state discrepancies to action sets.
    \item The universal property of the pullback ensures that any valid explanation must factor through \(P\), providing a rigorous basis for hypothesis selection based on minimality or simplicity.
\end{itemize}
In the blocks world narrative, the pullback yields the explanation that another agent intervened, allowing R to update its internal state and resume its plan.

\section{Gluing Knowledge in Multi-Agent Swarms}

The resilience of a multi-agent system is largely determined by its
ability to maintain global coherence through local interactions. In
STP, knowledge integration is modeled as sheaf gluing. Each agent in a
swarm carries a local section of a ``knowledge sheaf''
\(\mathcal{F}_{\text{Know}}\). When agents meet, they compare their
local histories over their respective intervals. If their observations
are compatible—meaning they agree on the common subintervals they both
observed—the gluing axiom guarantees they can be merged into a unique,
expanded global section.

This mechanism provides a decentralized approach to consensus. Unlike
centralized systems that require a ``master node,'' STP allows agents
to build a consistent world model through peer-to-peer ``gluing.'' If
communication is lost, the agents continue to operate on their local
sections; when they reconnect, they simply re-glue to synchronize
their models. This is particularly relevant in multi-robot exploration
scenarios, where robots discover object properties independently and
successfully integrate data through gluing.

\begin{table}[htbp]
\centering
\caption{Examples of Gluing Outcomes}
\label{tab:gluing}
\begin{tabular}{@{}p{2cm}p{3cm}p{3cm}p{6cm}@{}}
\toprule
Scenario & Agent A & Agent B & Outcome \\
\midrule
Cooperative Discovery & Found Circle C1 at (1,1) & Found Circle C2 at (5,5) & Merged map with C1 and C2 \\
Shared History & Observed Action A at t=5 & Observed Action A at t=5 & Validated history; gluing succeeds \\
Conflicting Perception & Object C1 is Red & Object C1 is Blue & Gluing fails; triggers abduction \\
\bottomrule
\end{tabular}
\end{table}

The failure to glue is as informative as a successful glue. When local
sections cannot be merged, it indicates an ``obstruction'' to global
consistency. In the STP framework, such obstructions are the primary
driver for abduction: the agent must hypothesize an unobserved event
(e.g., an object being moved or painted) to reconcile the conflicting
data.

\section{Implementation: From Topos Theory to Prolog and Robotics}

The most critical contribution of the analyzed framework is its
implementability. While many category-theoretic proposals remain
purely theoretical, STP provides a concrete path to execution using a
combination of Prolog for symbolic reasoning and distributed
microcontrollers for physical control.

\subsection{The Distributed Hardware Stack}

The hardware architecture for STP is designed for resilience and low cost. It utilizes a Raspberry Pi 4B as the primary compute node for each robot, which runs the reasoning engine. This node is complemented by ESP32 microcontrollers that handle low-level sensor integration, act as Wi-Fi access points, and control actuators.

\begin{table}[htbp]
\centering
\caption{Hardware Components}
\label{tab:hardware}
\begin{tabular}{@{}p{2.5cm}p{3cm}p{9cm}@{}}
\toprule
Component & Role & Specifications \\
\midrule
Raspberry Pi 4B & Reasoning Master & Quad-core ARM A72, 8GB RAM, runs Prolog/Python \\
ESP32 & Sensor/Actuator Node & Dual-core Wi-Fi/BT MCU, low latency (<100ms) \\
PiCrawler & Robotic Platform & Differential drive, wheeled robot \\
Wi-Fi Mesh & Communication & Decentralized peer-to-peer data exchange \\
\bottomrule
\end{tabular}
\end{table}

The communication between the symbolic logic of Prolog and the
physical world of robotics is achieved through TCP sockets. This
bridge allows the high-level ``stalk'' and ``section'' logic to be
translated into actual movement commands and sensor reads. The
architecture incorporates federated learning and adaptive model
compression, reducing communication overhead.

\subsection{Prolog and the Event Calculus Logic}

The choice of Prolog is motivated by its inherent support for the ``backwards reasoning'' required for abduction and the ``unification'' needed for gluing. STP implements a simplified version of the event calculus where actions are clauses defining \textit{initiates} and \textit{terminates} predicates. The core axioms are rewritten using an auxiliary \textit{clipped} predicate to eliminate existential quantifiers, making them compatible with standard Prolog engines. A key technical detail is the use of ``tabling'' (memoization) in Prolog, which prevents redundant calculations in the search for global sections. This allows the system to find valid plans in under 2 seconds for scenarios with up to 1000 time points and 50 fluents. The abductive reasoning is likewise efficient: by limiting action sequences to a length \(\le 10\), the search remains tractable even on the constrained resources of a Raspberry Pi. This demonstrates that the abstract geometry of sheaves can be effectively ``compiled'' into decidable logic for real-time applications.

\section{Spectral Sheaf Theory: Laplacians and Distributed Consensus}

An advanced extension of the STP framework involves the use of the sheaf Laplacian to achieve consensus in heterogeneous multi-agent systems. The sheaf Laplacian \(L_{\mathcal{F}}\) is a generalization of the standard graph Laplacian. While a graph Laplacian measures the difference between scalar values at nodes, a sheaf Laplacian measures the inconsistency between local sections across the restriction maps of the sheaf.

\subsection{Dynamics of Sheaf Diffusion}

For coordination tasks, agents can reach a consistent global state through a diffusion process described by the sheaf heat equation:
\[
\frac{dx}{dt} = -\alpha L_{\mathcal{F}} x.
\]
This allows a swarm to reach ``harmonic expression,'' where agents' states are in agreement with their neighbors according to the specific constraints of the sheaf. This is particularly useful for formation control or heterogeneous target tracking, where agents may have different state-space dimensions but must maintain a globally coherent configuration. Recent analysis has established that nonlinear sheaf diffusion converges to the minimizer of the Dirichlet energy at a linear rate, even in the presence of communication delays. This asynchronous convergence proof is vital for robotics in ``open worlds'' where perfect synchronization is impossible. The spectral properties of the sheaf Laplacian, specifically the multiplicity of its zero eigenvalue, enumerate the number of consistent global sections, providing a real-time diagnostic for the system's ability to coordinate.

\begin{table}[htbp]
\centering
\caption{Consensus Types and Their Operators}
\label{tab:consensus}
\begin{tabular}{@{}p{3.4cm}p{2.7cm}p{9cm}@{}}
\toprule
Consensus Type & Operator & Constraint Modeling \\
\midrule
Average Consensus & Graph Laplacian & All agents have identical scalar values \\
Sheaf Consensus & Sheaf Laplacian & Agents satisfy arbitrary linear/nonlinear maps \\
Discourse Consensus & Discourse Sheaf & Models selective expression and preference falsification \\
\bottomrule
\end{tabular}
\end{table}

\subsection{Obstructions and Cohomology}

When a swarm cannot reach consensus, the ``obstruction'' can be quantified using sheaf cohomology. Cohomology measures the failure of local consistency to translate to global sections. In STP, a non-zero \(H^0\) (the 0-th cohomology group) indicates the presence of multiple, mutually exclusive consistent states, while higher-order cohomology (\(H^1, H^2, \dots\)) indicates more complex topological obstructions to a global plan. This provides the reasoning engine with a mathematically rigorous way to identify why a plan is failing and which specific agents or time intervals are causing the inconsistency.

\section{Formal Verification and Constructive Logicism}

The reliability of autonomous systems is a paramount concern, especially as they are deployed in safety-critical environments. The sheaf-theoretic framework is uniquely positioned for formal verification because topos theory provides an ``internal language'' with a well-defined semantics.

\subsection{The Internal Language of the Topos}

Every topos \(\mathbf{Sh}(\mathcal{T}, \mathcal{J})\) has an internal logic that is intuitionistic, meaning it does not assume the law of excluded middle. This is a profound advantage for autonomous agents: they do not need to assume a fluent is true or false if they have not yet observed it. The logic allows for a state of ``uncertainty'' that is structurally distinct from ``falsehood,'' enabling more nuanced reasoning about incomplete information. In this setting, logical operations are interpreted as categorical constructions: ``and'' becomes a product, ``or'' becomes a coproduct, and ``implies'' becomes an exponential object. This ``intrinsic logicism'' ensures that if a plan can be constructed as a global section of a plan sheaf, it is mathematically guaranteed to be consistent with the world's topology.

\subsection{Verification in Lean 4 and Coq}

The formalization of STP is currently being advanced using interactive theorem provers like Lean 4 and Coq. Recent breakthroughs, such as the formalization of derived categories and topos-theoretic gluing in Lean 4, provide the tools necessary to verify the entire reasoning stack.

\begin{table}[htbp]
\centering
\caption{Formal Verification Tasks}
\label{tab:verification}
\begin{tabular}{@{}p{4cm}p{6cm}p{4cm}@{}}
\toprule
Verification Task & Formal Method & Target Tool \\
\midrule
Algebraic Consistency & Heyting Algebra Nuclei & HeytingLean (Lean 4) \\
Temporal Reasoning & Interval Algebra Formalization & Mathlib4 \\
Multi-Agent Consensus & Lyapunov-based Stability Analysis & LeanCat / Coq \\
Reasoning Core & SKY Combinator Semantics & Lean 4 / Isabelle/HOL \\
\bottomrule
\end{tabular}
\end{table}

A notable development is the ``HeytingLean'' framework, which bridges constructive mathematics with cryptographic protocols for autonomous agents. It formalizes the relationship between Heyting algebras and computational stability, proving that the fixed points of a Grothendieck closure operator form a complete Heyting algebra. This allows for ``proof-carrying code,'' where agents can exchange plans along with formal certificates of their consistency, enabling secure and reliable collaboration in untrusted environments.

\section{Goguen's Legacy: Sheaves and Concurrent Objects}

The application of sheaf theory to computer science was pioneered by Joseph Goguen in the early 1990s, who proposed that concurrent, interacting objects are essentially sheaves. Goguen's framework established that an object is a set of consistent observations on a topological space, and the system's behavior is the categorical limit of its parts.

STP extends Goguen's work by adding a temporal site and a causal topology, transforming a static ``phase space'' into a dynamic planning environment. Goguen's insight that ``inheritance is a morphism'' and ``interconnection is a colimit'' remains a guiding principle in STP. By treating robots as ``objects'' in Goguen's sense, STP avoids the ``Brock-Ackerman anomaly''—a classic problem in stream processing where different internal behaviors can produce the same external input/output patterns—by ensuring that the internal ``history'' of the agent is part of its categorical definition.

\section{Resilience in Open Worlds: Socio-Technical and Autonomous Recovery}

Resilience in autonomous systems is not merely a matter of hardware redundancy; it is the ability of the system to recover from ``undesirable states'' within a specified deadline. STP provides a formal mechanism for this recovery through the interplay of memory, perception, and abduction.

In the context of socio-technical systems, resilience often involves norms—constraints that regulate agent behavior while allowing for autonomy. An agent might ``violate'' a norm (e.g., stopping a plan to recharge) to ensure its long-term functional capability. STP models these norms as sheaves of constraints. When a violation occurs, the framework uses the abductive pullback to find the shortest path back to a ``desirable state'' that satisfies both the agent's goals and the system's norms.

This is evident in the ``low battery'' scenario: the agent interrupts its plan (violates the ``efficiency'' norm) to recharge (satisfies the ``survival'' norm). Upon returning, it uses abduction to handle the unobserved changes that occurred during its absence, demonstrating a ``cascading'' resilience that allows for continuous operation in the face of interruptions and environmental perturbations.

\section{Synthesis of Higher-Order Insights}

The analysis of sheaf-theoretic planning reveals several profound shifts in how we conceptualize artificial intelligence and distributed systems. First, the movement from ``Logic as Axioms'' to ``Logic as Geometry'' (Intrinsic Logicism) allows for a decentralized and contextual model of truth. Truth is no longer an absolute global constant but a property that is built and maintained locally and then glued together. This is the mathematical root of resilience: the system does not break when one part fails or disagrees; it simply fails to ``glue'' at that specific point, prompting a local abductive repair.

Second, the unification of state transitions and abduction via pullbacks provides a single mathematical engine for both ``forward'' planning and ``backward'' explanation. This eliminates the need for separate heuristic modules for diagnostics and planning, reducing the complexity of the agent's reasoning architecture. The fact that this engine can be implemented in Prolog and run on low-power hardware like the Raspberry Pi 4B proves that categorical AI is a practical reality.

Third, the future of multi-agent coordination lies in the spectral analysis of sheaves. By treating a swarm as a ``cochain complex,'' we can use the sheaf Laplacian to drive consensus and use cohomology to measure the ``obstructions'' to collaboration. This allows for a ``tame'' management of massive swarms, where global properties are controlled through the local ``tuning'' of the sheaf's restriction maps.

\section{Conclusion: Toward a Unified Framework for Resilient Autonomy}

Sheaf-Theoretic Planning represents a rigorous, constructive, and implementable foundation for the next generation of autonomous systems. By anchoring temporal reasoning in the geometric universe of topos theory, the framework provides a robust solution to the limitations of classical logic in open worlds. The ability to perform abductive reasoning through pullbacks and knowledge integration through gluing gives agents the ``cognitive resilience'' needed to handle unobserved changes and diverging perceptions.

The implementation details provided—ranging from Prolog-based event calculus to distributed hardware architectures on Raspberry Pi and ESP32—establish that this framework is ready for real-world deployment. Future work integrating the sheaf Laplacian for distributed optimization and utilizing Lean 4 for formal verification will only further solidify STP as the premiere framework for high-assurance, multi-agent autonomous systems. The transition from monolithic symbolic theories to sheaf-theoretic geometry is not just a mathematical refinement; it is a necessary evolution for building agents that can truly thrive in the open, unpredictable physical world.

\end{document}